\documentclass{article}

\usepackage[final]{neurips_2022}

\usepackage{natbib} % has a nice set of citation styles and commands
\usepackage[utf8]{inputenc} % allow utf-8 input
\usepackage[T1]{fontenc}    % use 8-bit T1 fonts
\usepackage{hyperref}       % hyperlinks
\usepackage{url}            % simple URL typesetting
\usepackage{booktabs}       % professional-quality tables
\usepackage{amsfonts}       % blackboard math symbols
\usepackage{nicefrac}       % compact symbols for 1/2, etc.
\usepackage{microtype}      % microtypography
\usepackage{xcolor}         % colors

\usepackage{multirow}
\usepackage{mathtools}
\DeclareMathOperator*{\argmax}{arg\,max}

\title{Gumbel-Softmax Selective Networks}

\author{%
  Mahmoud Salem$^{1,2,3}$\thanks{This work was done during an internship at Borealis AI.} \quad Mohamed Osama Ahmed$^1$ \quad Frederick Tung$^1$ \quad Gabriel Oliveira$^1$ \\
  $^1$Borealis AI \quad $^2$Vector Institute \quad $^3$University of Guelph\\
  \texttt{mahmoud.gemy18@gmail.com} \\
  \texttt{\{mohamed.o.ahmed, frederick.tung, gabriel.oliveira\}@borealisai.com} \\
}

\begin{document}

\maketitle

\begin{abstract}
  ML models often operate within the context of a larger system that can adapt its response when the ML model is uncertain, such as falling back on safe defaults or a human in the loop. This commonly encountered operational context calls for principled techniques for training ML models with the option to abstain from predicting when uncertain. Selective neural networks are trained with an integrated option to abstain, allowing them to learn to recognize and optimize for the subset of the data distribution for which confident predictions can be made. However, optimizing selective networks is challenging due to the non-differentiability of the binary selection function (the discrete decision of whether to predict or abstain). This paper presents a general method for training selective networks that leverages the Gumbel-softmax reparameterization trick to enable selection within an end-to-end differentiable training framework. Experiments on public datasets demonstrate the potential of Gumbel-softmax selective networks for selective regression and classification.
\end{abstract}

\section{Introduction}\label{sec:intro}

When an ML model is uncertain about its prediction, for example due to the uniqueness of the input with respect to previously observed training samples, it is often preferable for the model to abstain from making a prediction, instead of making a poor prediction that could erode user confidence or lead to harmful downstream consequences. In cases of abstention, the system may fall back on expert judgment or safe defaults. The automatic learning of an abstention policy frees ML system developers from having to hand-craft a set of selection rules based on heuristics.

\textit{Selective networks} are trained with an integrated reject option, i.e., the option to abstain from making a prediction when the model is uncertain \citep{geifman2019selectivenet}.
Optimizing selective networks is challenging because of the non-differentiability of the binary selection operation (the decision of whether to select or abstain). In the conventional formulation of selective networks, the non-differentiability of selection is handled by replacing the binary selection operation with a soft relaxation. However, this approximation means that in practice the selective network does not perform selection during training, but instead assigns a soft instance weight to each training sample. 

In this paper, we present Gumbel-softmax selective networks, which enable binary selection decisions during training while preserving end-to-end differentiability using the Gumbel-softmax reparameterization trick \citep{jang2017,maddison2017}. 
The proposed technique for training selective networks is general and does not assume a particular prediction task (e.g. classification). It leverages a principled tool to perform selection or abstention within an end-to-end training framework. Experiments on four public datasets demonstrate the potential of Gumbel-softmax selective networks for both selective regression and selective classification tasks.

\section{Related Work}

In practice, it is often useful for an ML system to have the option of abstaining from making a prediction when it detects a situation of high uncertainty. Given that the system has the option to abstain, an important question to ask is how we can train the model \textit{with the knowledge that it is allowed to abstain}. By integrating this option into model training, the model can learn to automatically recognize and optimize for the part of the data distribution for which confident predictions can be made, instead of attempting to fit the entire data distribution at training time and applying hand-crafted abstention rules at inference time.

How to train a neural network with the knowledge that it is allowed to abstain has received relatively little attention in the ML community. 
\citet{geifman2019selectivenet} proposed the modern selective network (SelectiveNet), which adds a dedicated selection head to the base network. The network is trained to optimize the task performance criterion, such as classification accuracy, given a target level of coverage: the proportion of input samples for which the network should make predictions. %For example, a target coverage of 90\% means that the network should abstain at most 10\% of the time. We discuss the training process in more detail in Section \ref{subsec:selectivenet}.
\citet{Liu2019deepgamblers} proposed to add the abstention option as a separate class that can be predicted. A threshold is applied to the score of the abstention class to achieve a desired level of coverage without re-training. However, this approach can be applied to classification networks only. We propose a general approach that can be applied to any predictive task.
\citet{huang2020selfadaptivetraining} used the selective classification task to illustrate the potential of their self-adaptive training technique, which improves generalization performance in the presence of noisy training data. %Our contribution is orthogonal and can be combined with self-adaptive training.

\section{Method}
\subsection{Preliminaries: Selective Networks}
\label{subsec:selectivenet}

A \textit{selective neural network} can be defined as a pair $(f,g)$, where $f$ is a prediction function and $g$ is a binary selection function, such that the output of the network is given by \citep{geifman2019selectivenet}:
\begin{equation}
(f,g)(x) = 
\begin{cases}
 f(x) & \text{ if } g(x)=1 \\ 
 \text{Abstain}& \text{ if } g(x)=0.
\end{cases}
\end{equation}

Selective networks trade off prediction performance against coverage: the proportion of input samples that the network selects (i.e., makes predictions for). Given a set of $m$ training data points $\{{(x_i, y_i)}\}_{i=1}^{m}$, the empirical coverage is defined as
\begin{equation}
\widehat{\phi}(g) = \frac{1}{m} \sum_{i=1}^{m} g(x_i) \, ,
\end{equation}
and the empirical selective risk is defined as 
\begin{equation}
\widehat{r}(f,g) = \frac{\frac{1}{m} \sum_{i=1}^{m} \ell(f(x_i),y_i)g(x_i)}{\widehat{\phi}(g)} \, ,
\end{equation}
where $\ell$ is a loss function such as cross-entropy for classification or mean squared error for regression.
The overall training objective is then a weighted combination of the empirical selective risk and a penalty term that penalizes differences between the empirical coverage and a pre-specified target coverage:
\begin{equation}
\label{eq:Lfg}
\mathcal{L}_{(f,g)} = \widehat{r}(f,g) + \lambda \Psi (c - \widehat{\phi}(g)) \, ,
\end{equation}
where $c$ is a pre-specified target coverage, $\Psi$ is a penalty function (e.g. $\Psi(a) = max(0,a)^2$), and $\lambda$ is a balancing hyperparameter.

Optimizing Eq.~\ref{eq:Lfg} is challenging because of the non-differentiability of the binary selection function $g$. \citet{geifman2019selectivenet} handle the non-differentiability of selection by replacing the binary function $g$ with a relaxed function $g : \mathcal{X} \to [0, 1]$. While this addresses the differentiability issue, the approximation means that in practice the selective network does not perform selection during training, but instead assigns a soft instance weight to each training sample.
This introduces a gap between training and inference.
%This is not aligned with the goal of the optimization process which aims at minimizing the loss over only the selected examples to achieve a desired coverage. 
To address this discrepancy, in the following we describe a differentiable method for enabling binary selection during training while preserving end-to-end training using the Gumbel-softmax reparameterization trick.

\subsection{Gumbel-softmax Selective Networks}

\begin{figure}[t]
  \centering
  \includegraphics[width=0.7\linewidth]{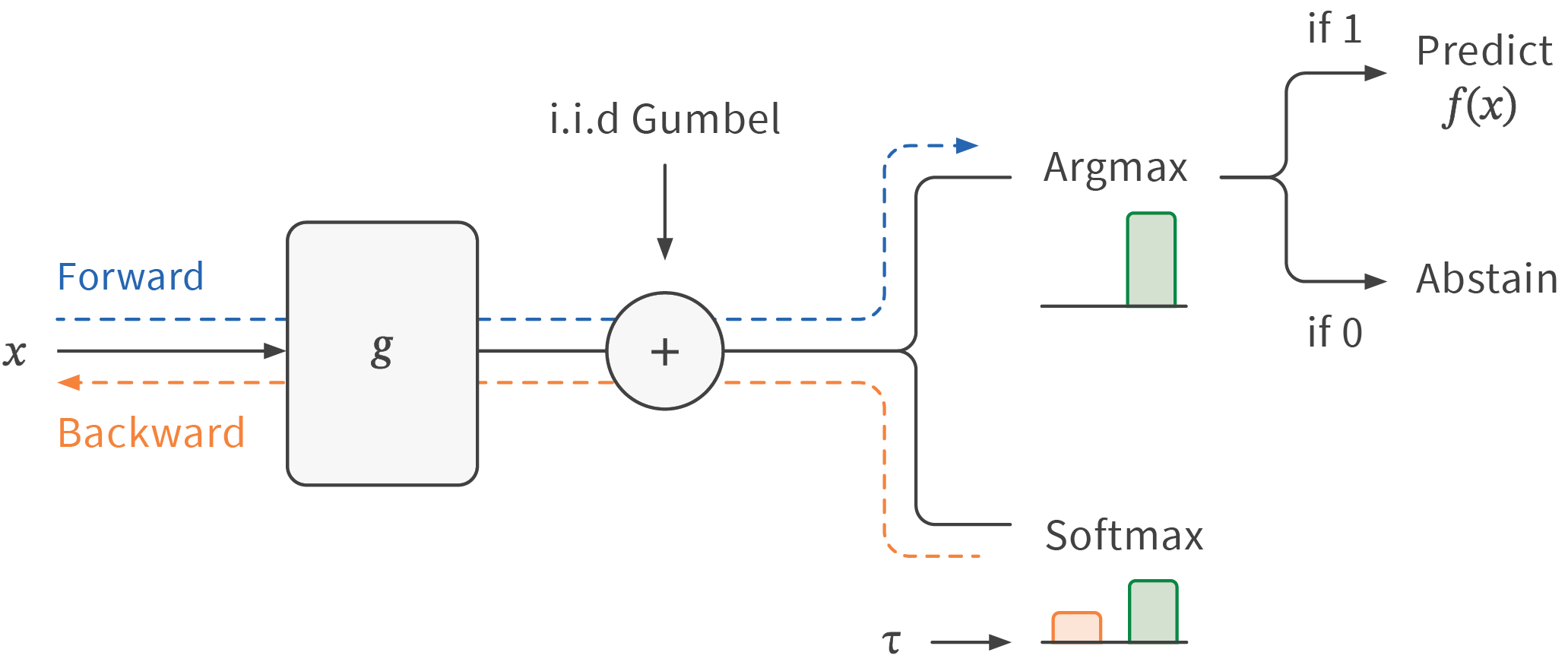}
  \caption{Gumbel-softmax selective networks leverage the Gumbel-softmax reparameterization trick \citep{jang2017,maddison2017} to enable selection (abstention) decisions within an end-to-end differentiable training framework. The temperature parameter $\tau$ is annealed over time such that the softmax approaches the argmax.}
  \label{fig:summary}
\end{figure}

The reparameterization trick \citep{kingma2014,rezende2014} in deep learning allows us to replace a stochastic computation graph by a differentiable computation graph with learnable parameters, acting on noise from a fixed base distribution. For example, suppose we want a stochastic node in a neural network that performs sampling from a normal distribution parameterized by mean $\mu$ and standard deviation $\sigma$. We cannot backpropagate through this stochastic node because of the non-differentiability of the sampling operation. However, we can replace this stochastic node with a parameterized differentiable computation that takes noise as input: the computation takes input noise sampled from the standard normal $\mathcal{N}(0, 1)$, scales it by $\sigma$, and then shifts the result by $\mu$. Since $\mu$ and $\sigma$ can be generated by deterministic neural network layers trainable by backpropagation, this reparameterization effectively enables sampling from an arbitrary, learnable normal distribution. %To reiterate, we moved the dependency on parameters $\mu$ and $\sigma$ from the stochastic computation graph to a differentiable computation graph that acts on base noise.

We now revisit the conventional selective network formulation and show how we can use the reparameterization trick to perform binary selection while preserving end-to-end training.
Let us re-define the output of $g$ as the probability of selecting the input (i.e., the probability the network should make the prediction instead of abstaining). The selection function becomes a stochastic operator that selects the input with probability $g$. Similar to the example at the beginning of this subsection, we have a stochastic node that performs a sampling operation. However, instead of sampling from a normal distribution, we want to sample from the Bernoulli distribution, $Bernoulli(g)$.

The Gumbel-softmax reparameterization trick \citep{jang2017,maddison2017} allows us to reparametrize a stochastic node that samples from a categorical distribution, again by replacing it with a differentiable function of learnable parameters, acting on noise from a base distribution. Given a categorical distribution of $k$ events with probability $\pi_1, ..., \pi_k$, we compute $\log \pi_1, ..., \log \pi_k$, and to each of these terms we add i.i.d. noise sampled from the Gumbel distribution \citep{gumbel1954}. We can then draw a stochastic sample $z$ (represented by a one-hot vector) by taking the argmax:
\begin{equation}
    z = one\_hot (\argmax_i [G_i + \log \pi_i]) \, ,
\end{equation}
where $G_i \sim Gumbel(0, 1)$. To allow end-to-end training, we approximate the argmax with a softmax, which gives a softened vector $\tilde{z}$:
\begin{equation}
\tilde{z}_i = \frac{\exp \, ((\,\log\pi_i+G_i))/\tau)}{{\sum}_{j=1}^k{\exp \, ((\,\log\pi_j+G_j)/\tau)}} \, , \quad \text{for } i=1,...,k
\end{equation}
The temperature parameter $\tau > 0$ determines the sharpness of the softmax, and is annealed over time towards zero to recover the argmax. As $\tau \to \infty$, the Gumbel-softmax distribution converges to the uniform distribution, and as $\tau \to 0$, the Gumbel-softmax distribution converges to the categorical distribution. Therefore, we have moved the dependency on parameters $\pi_1, ..., \pi_k$ from the non-differentiable stochastic sampling function to a differentiable function consisting of softmax and log operations acting on base noise, which can be trained end-to-end with backpropagation.

Putting it all together, we perform binary selection by applying the Gumbel-softmax reparameterization trick with $\pi_1=g, \pi_2=1-g$. In the forward pass, we use the argmax form to perform binary selection. In the backward pass, we use the softmax form with temperature annealing to compute gradients and enable end-to-end training. %This variation of using argmax in the forward pass and softmax in the backward pass is also known as the straight-through Gumbel-softmax \citep{jang2017}. 
Figure \ref{fig:summary} shows a visual summary of the proposed approach.

\section{Experiments}

In this section, we demonstrate the potential of Gumbel-softmax selective networks on four public datasets. Due to space limitations, we defer dataset and implementation details to the supplementary.
Selective networks trained at the same level of target coverage may differ in the actual coverage achieved in evaluation (i.e., the number of predictions made on the test set). %due to distribution shift or random train-test variations. 
For a fair comparison, we apply coverage calibration \citep{geifman2019selectivenet} to equalize the number of test predictions across all approaches. For example, when evaluating at a coverage level of 70\%, we compute the error metrics over the 70\% most confident predictions (highest $g$ values) among the test samples.

% \subsection{Selective Regression Results}

Table \ref{tab:regression_results} summarizes the experimental results for Gumbel-softmax selective networks and SelectiveNets on three public regression datasets, averaged over five trials.
We train all models from scratch, and for a fair comparison all shared hyperparameters and train budgets are the same. 
%We report regression error metrics for coverages ranging from 100\% to 50\%.
On the Concrete Compressive Strength dataset, the results we obtain for SelectiveNet are better than those reported in the original paper \citep{geifman2019selectivenet} as we found that applying a learning rate decay schedule, instead of a constant learning rate as in \citet{geifman2019selectivenet}, substantially boosts performance.
Gumbel-softmax selective networks consistently outperform SelectiveNets at every coverage level on all three regression datasets.

\begin{table}
\centering
\scriptsize
\caption{Selective regression results on Concrete Compressive Strength, California Housing, and Ames Housing datasets. For the housing datasets, errors are computed in units of \$10,000. We highlight in bold the lowest error rates.}
\begin{tabular}{c|cc|cc|cc}
\toprule[1.2pt]
\multirow{3}{*}{Coverage} & \multicolumn{2}{c|}{Concrete Compressive Strength} & \multicolumn{2}{c|}{California Housing} & \multicolumn{2}{c}{Ames Housing} \\
& \multicolumn{2}{c|}{MSE ($\downarrow$)} & \multicolumn{2}{c|}{MAE (10,000's, $\downarrow$)} & \multicolumn{2}{c}{MAE (10,000's, $\downarrow$)} \\
& Gumbel-softmax & SelectiveNet & Gumbel-softmax & SelectiveNet & Gumbel-softmax & SelectiveNet \\
\midrule \midrule
100 & 32.84$\pm$2.50 & 32.82$\pm$0.67 & 4.51$\pm$0.03 & 4.55$\pm$0.05   & 1.68$\pm$0.07 & 1.64$\pm$0.04 \\
90 & \textbf{25.13$\pm$1.22} & 26.56$\pm$2.82 & \textbf{4.19$\pm$0.05} & 4.36$\pm$0.11 & \textbf{1.22$\pm$0.04} & 1.25$\pm$0.05 \\
80 & \textbf{21.15$\pm$0.83} & 21.80$\pm$3.25 & \textbf{3.92$\pm$0.07}  & 4.24$\pm$0.17 & \textbf{1.10$\pm$0.05}  & 1.11$\pm$0.03 \\
70 & \textbf{16.17$\pm$1.85} & 18.59$\pm$2.50  & \textbf{3.66$\pm$0.04} & 3.97$\pm$0.18 & \textbf{1.04$\pm$0.01} & 1.07$\pm$0.03 \\
60 & \textbf{13.72$\pm$2.44} & 17.59$\pm$2.23 & \textbf{3.38$\pm$0.09} & 3.99$\pm$0.23 & \textbf{0.97$\pm$0.03} & 1.00$\pm$0.04 \\
50 & \textbf{11.15$\pm$2.11} & 14.43$\pm$2.57 & \textbf{3.22$\pm$0.15} & 3.78$\pm$0.15 & \textbf{0.95$\pm$0.06} & 1.01$\pm$0.05 \\
\bottomrule[1.2pt]
\end{tabular}
\label{tab:regression_results}
\end{table}

Following \cite{feng2022}, Table \ref{tab:classification_results} summarizes the experimental results on the ImageNet-100 dataset, averaged over five trials.  Gumbel-softmax selective networks modestly outperform SelectiveNets at higher coverage levels; both methods perform comparably at lower coverage levels.

\begin{table}
\centering
\footnotesize
\caption{Selective classification results on ImageNet-100. We highlight in bold the lowest error rates.}
\begin{tabular}{c|cc}
\toprule[1.2pt]
\multirow{3}{*}{Coverage} & \multicolumn{2}{c}{ImageNet-100} \\
& \multicolumn{2}{c}{Top-1 Accuracy ($\uparrow$)} \\
& Gumbel-softmax & SelectiveNet \\
\midrule \midrule
100 & 86.16$\pm$0.15 & 86.07$\pm$0.11 \\
90 & \textbf{89.76$\pm$0.64} & 88.68$\pm$0.30 \\
80 & \textbf{93.33$\pm$0.47} & 92.59$\pm$0.18 \\
70 & \textbf{96.03$\pm$0.33} & 95.86$\pm$0.45  \\
60 & 97.79$\pm$0.34 & \textbf{97.83$\pm$0.28} \\
50 & \textbf{99.12$\pm$0.49} & 99.06$\pm$0.23  \\
\bottomrule[1.2pt]
\end{tabular}
\label{tab:classification_results}
\end{table}

\section{Conclusion}

ML models are often deployed not in isolation, but as part of a larger system, with non-ML logic, legacy processes, or humans in the loop. In operational contexts where the system has the option of falling back on supporting processes when the ML model is uncertain, the option to abstain should be integrated directly in the ML model training. We hope that our ideas on how to train selective networks will reinvigorate interest in this practical problem.

\bibliography{neurips_2022}

\newpage
\appendix

\begin{center}
\Large{\textbf{Supplementary Material}}
\end{center}

\section{Datasets}

\textbf{Concrete Compressive Strength} \citep{concretecompressivestrength} is a regression dataset from the UCI Machine Learning Repository \citep{uci} that is used in the experimental evaluation of SelectiveNet \citep{geifman2019selectivenet}. It consists of 1,030 instances and the task is to predict the compressive strength given eight numerical input variables. As there is no standard training-testing split, we randomly split the dataset into 60\% for training, 20\% for held-out validation, and 20\% for testing. After tuning hyperparameters on the validation set, we trained the final models on the combined training-validation set and generated the results on the testing set.

\textbf{California Housing} \citep{calihousing} is a regression dataset. It consists of 20,640 instances and the task is to predict median housing values of California districts given eight input features. As there is no standard training-testing split, we randomly split the dataset into 80\% for training (16,512 instances) and 20\% for testing (4,128 instances). For hyperparameter searching purposes, we further divide the training set into 80\% training and 20\% validation. After hyperparamater exploration, the combined training-validation set is used to train the final models for evaluation on the testing set.

\textbf{Ames Housing} \citep{ames_kagle} is a house price regression dataset featuring houses sold in Ames, Iowa during the period from 2006 to 2010. 
The dataset has 1,460 instances and the goal is to predict the sale price of the house. The dataset includes 79 features divided into categorical and numerical. Based on available resources we dropped columns with more than 80\% of its samples missing, which are Alley, PoolQC, MiscFeature and Fence. GarageYrBlt is also removed due to high redundancy to the MasVnrArea feature.  
The training set contains 1,022 instances and the testing set contains 438 instances. For hyperparameter searching purposes we further divide the training set into 70\% training and 30\% validation. After hyperparamater exploration the entire 1022/438 training/testing set is used to generate the final results. The dataset contains a number of missing values in both its numerical and categorical features. In order to replace the missing values we perform mean value imputation along each numerical column and most frequent value for each categorical column. Additionally, categorical data was also converted to one-hot encoding representation to obtain the final configuration used during experiments.

\textbf{ImageNet-100} \citep{tian2019} is a 100-class subset of ImageNet. Details on its construction can be found in the supplementary of \citet{tian2019}.

\section{Implementation Details}

We follow the recommendation in \citet{geifman2019selectivenet} and use an auxiliary prediction head as a regularizer during training. The auxiliary head is discarded after training and there is no additional overhead at inference time.

In the regression experiments, we adopted multilayer perceptron (MLP) backbones. For the Concrete Compressive Strength (CCS) dataset, we utilize a single hidden layer MLP with 64 neurons with ReLU and batch normalization, following the same setting from \citet{geifman2019selectivenet}. The California Housing dataset backbone is composed of a MLP with 2 hidden layers of 100 neurons each with ReLU. For the Ames Housing dataset, we use a two hidden layer MLP with 100 neurons with ReLU and batch normalization. The networks were trained for 800 epochs for the CCS and Ames datasets and for 1000 epochs for the California Housing dataset. All datasets used adam as optimizer, with initial learning rate of 0.007 and decay at epochs 400, 500, 600, 700 with a factor of 0.5 for the CCS dataset, an initial learning rate of 0.007 and decay at epochs 250, 500, 750  with a factor of 0.1 for the California Housing dataset, and an initial learning rate of 0.007 and decay at epochs 150, 250 with a factor of 0.1 for the Ames Housing dataset. The Gumbel-softmax temperature $\tau$ was initialized to 30 and annealed using multi-step decay by the rate of 0.985 every 5 epochs for the Concrete Compressive Strength and California Housing datasets. Ames Housing dataset used an initial $\tau$ of 10 and annealed it using multi-step decay by the rate of 0.995 every 5 epochs.

In the classification experiments, we used the ResNet-34 architecture proposed by \citet{resnet16}. Standard data augmentation was used in all classification experiments consisting of horizontal flips, vertical and  horizontal shifts and rotations. We used stochastic gradient descent (SGD) for optimization with momentum 0.9 and starting with initial learning rate of of 0.1. We lengthened the training schedule by a factor of two, applying a learning rate decay of 0.5 every 100 epochs for a total of 600 epochs. The Gumbel-softmax temperature $\tau$ was initialized to 5 and annealed using multi-step decay by the rate of 0.985 every 5 epochs.

\section{Limitations}

In practice, selective networks often operate in the context of a larger system. They are only part of the solution towards deploying a robust ML system. Other components such as out-of-distribution detection, calibration, bias mitigation, and automatic detection of systematic errors (slice discovery), are also important.

%%%%%%%%%%%%%%%%%%%%%%%%%%%%%%%%%%%%%%%%%%%%%%%%%%%%%%%%%%%%

\end{document}